\documentclass[11pt]{article}

\usepackage[numbers]{natbib}

\usepackage{graphicx}
\usepackage{amsmath,amssymb}
\usepackage{authblk}
\usepackage{url}
\usepackage{booktabs}
\usepackage{makecell}
\usepackage{geometry}
\usepackage{algorithm}
\usepackage{algpseudocode}
\usepackage{float}

\usepackage[table]{xcolor}
\usepackage{colortbl}
\definecolor{lightred}{RGB}{255, 230, 230}

\usepackage[colorlinks=true,linkcolor=blue,citecolor=blue,urlcolor=blue]{hyperref}

\title{Empowering Microscopic Traffic Simulators with Realistic Perception using Surrogate Sensor Models}

\author[1]{Tianheng Zhu}
\author[1,*]{Yiheng Feng}

\affil[1]{Lyles School of Civil and Construction Engineering, Purdue University, West Lafayette, Indiana, United States}
\affil[*]{Correspondence to: feng333@purdue.edu}

\date{} 

\begin{document}
\maketitle

\begin{abstract}
Simulation is central to the evaluation of intelligent transportation system (ITS) applications. As ITS increasingly incorporates autonomous vehicle (AV) technologies as fleet vehicles and/or mobile sensors, accurate modeling of their perception capabilities becomes essential in constructing high-fidelity simulation models. While game-engine-based simulators can mimic a realistic perception environment through 3D scene rendering and raw sensor data generation, they face scalability challenges in simulating transportation networks with a large number of AVs, due to substantial computational overhead. In contrast, microscopic traffic simulators can scale efficiently but lack perception modeling capabilities. To bridge this gap, we propose MIDAR, a surrogate LiDAR detection model that mimics realistic LiDAR detections using only high-level features, which can be easily obtained from microscopic traffic simulators. Specifically, MIDAR predicts true positives (TPs) and false negatives (FNs) from ideal LiDAR detection results based on the relative positions and dimensions of surrounding objects. 
To incorporate a physical prior on LiDAR visibility, MIDAR augments geometric features with a ray-hit signal that approximates LiDAR point coverage.
A Refined Multi-hop Line-of-Sight (RM-LoS) graph is constructed to encode the occlusion relationship between the ego AV and surrounding vehicles. MIDAR then employs a geometry-aware Graph Transformer to measure the impact of occlusion on ego AV perception along each RM-LoS chain. The proposed MIDAR model achieves an AUC of 0.94 in approximating LiDAR detection results with data collected from CARLA, an AV simulator, and an AUC of 0.86 with real-world LiDAR data collected from the nuScenes dataset. To demonstrate the necessity of the proposed surrogate sensor model, two ITS applications, cooperative-perception-based adaptive signal control and vehicle trajectory reconstruction, are integrated with MIDAR to further validate its realism and practicality. As shown in the results, MIDAR generates more realistic detection results as well as application-level performance metrics (e.g., vehicle delay, trajectory reconstruction accuracy) than simplified perception models. Meanwhile, MIDAR introduces very low computational overhead, which can be seamlessly integrated into large-scale, real-time traffic simulators without becoming a performance bottleneck.
The code and data are publicly available at \url{https://github.com/Purdue-CART-Lab/MIDAR}. 
\end{abstract}

\section{Introduction}
Over the past decade, autonomous vehicle (AV) technologies have become an increasingly integral part of intelligent transportation systems (ITS), reshaping the way transportation systems are operated and managed. Beyond transporting people and goods, AVs also function as mobile sensing platforms that generate valuable data for various ITS applications \cite{ye2019evaluating,garg2023can,peng2021connected,luo2024stabilizing,wang2023general}. For example, when multiple AVs share sensing data simultaneously, they create a cooperative perception (CP) environment and enable cooperative driving strategies to improve safety and mobility of both AV fleets and human driven vehicles \cite{chen2021graph, liu2024reinforcement}. From the transportation infrastructure perspective, adaptive traffic signal control systems can utilize AV observations through vehicle-to-infrastructure (V2I) communication to more accurately characterize approaching vehicle demand and improve intersection efficiency \cite{li2024a}. Researchers have also explored the use of such shared AV observations, usually partial due to low penetration rates, to reconstruct complete vehicle trajectories \cite{chen2024macro, zhang2024vehicle, zhu2026cptrajrecon} and estimate microscopic traffic states \cite{li2021cooperative} for traffic monitoring and control.


However, large-scale real-world experiments of AV-enabled ITS applications remain impractical due to substantial financial, technical, and operational constraints, as well as potential public safety risks. As a result, simulation has become the primary approach for prototyping and evaluating emerging AV technologies, which supports controlled experimentation at scale and reproducible results under diverse conditions.

\begin{figure}[!htbp]
    \centering
    \includegraphics[scale=0.6]{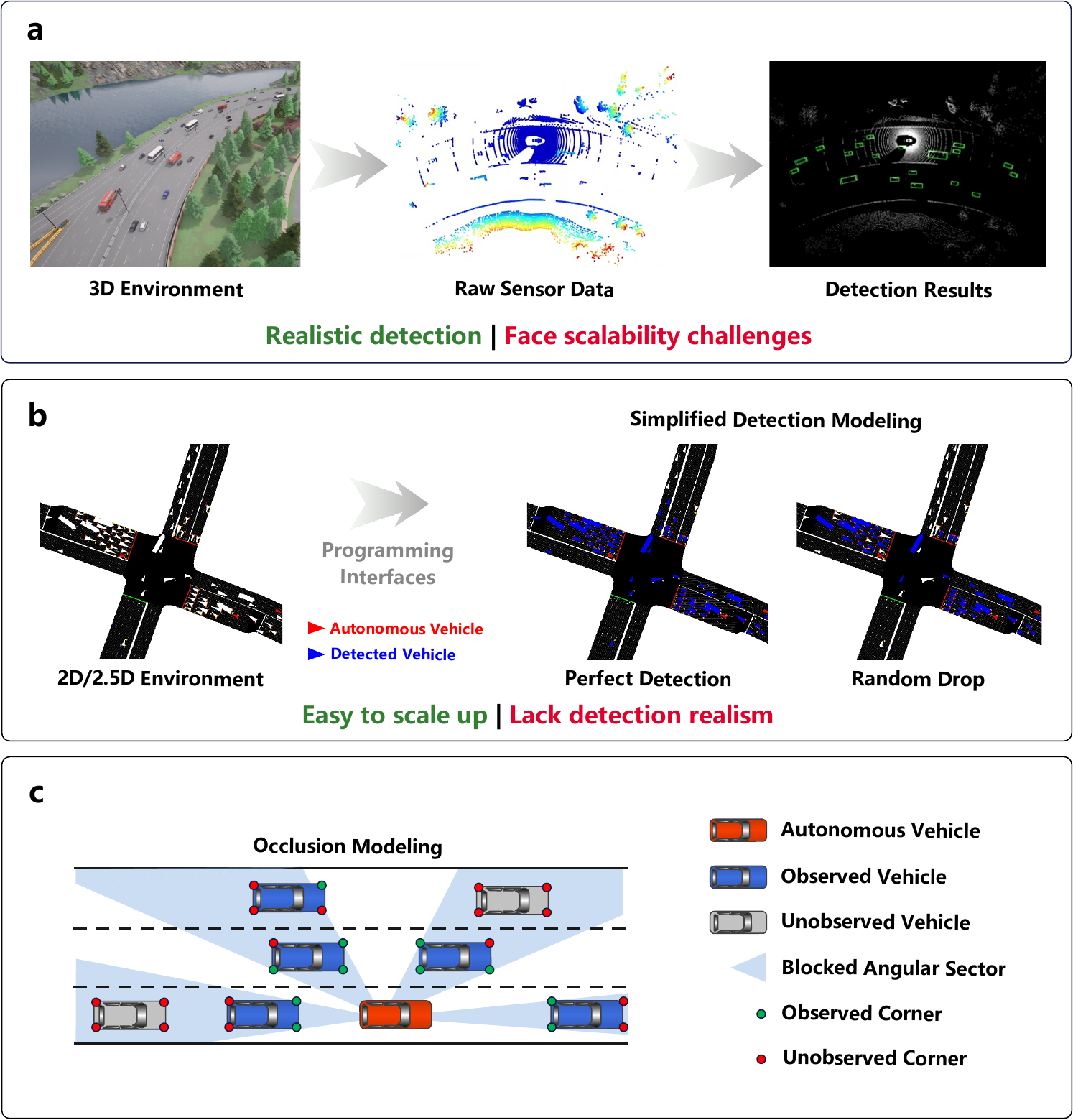}
    \caption{\textbf{Illustration of sensor data generation and detection in different types of simulators}
    \textbf{a.} Game-engine-based simulators provide raw sensor signals, to which object detection algorithms can be directly applied to produce detection results.
    \textbf{b.} Microscopic traffic simulators apply simplified detection models such as perfect detection and random drop.
    \textbf{c.} Rule-based approach for microscopic traffic simulators to incorporate geometry-based physical visibility.}
    \label{fig:simulators}
\end{figure}

In general, simulation platforms for AV-enabled ITS applications can be categorized into two types: game-engine-based simulators and microscopic traffic simulators. Recent advances in computer graphics and high-performance computing devices have made it possible for game-engine-based simulators to vividly approximate real-world driving environments using high-fidelity 3D rendering and physics-based vehicle dynamics. Representative platforms in this category include CARLA \cite{CARLA}, AWSIM \cite{AWSIM}, AirSim \cite{AirSim}, LGSVL \cite{LGSVL}, etc. However, these simulators focus on mimicking the autonomous driving pipeline (i.e., perception, planning, and control) from the ego vehicle’s perspective and are less suitable for ITS applications, which typically simulates a large number of multi-modal road users in a large traffic network. In contrast, microscopic traffic simulators trade fidelity for computational efficiency and scalability. The simulation environments are typically 2D or 2.5D and provide only vehicle-level features (e.g., vehicle position, dimension, and motion) without detailed perception or physics modeling. Vehicle dynamics are governed by simplified driving models such as car-following and lane-changing rules, enabling efficient and flexible simulation of traffic flow and multi-vehicle interactions. Well-known microscopic traffic simulators include SUMO \cite{krajzewicz2012recent}, VISSIM \cite{vissim}, etc. To combine the strengths of both types of simulators, co-simulation platforms, such as CDASim \cite{bayartsengel2024enhancing}  and OpenCDA \cite{xu2021opencda}, integrate traffic-flow models with high-fidelity environment generation, enabling more comprehensive evaluation of autonomous driving and AV-enabled ITS applications.

To simulate AVs in simulation environments, generating sensor data is a fundamental requirement. Game-engine-based simulators support flexible setup of multi-modality sensor suites and generation of raw sensor signals, like RGB camera images and LiDAR point clouds, as shown in Fig.~\ref{fig:simulators}a. Object detection algorithms can be directly applied to the simulated sensor data to generate detection results. These high-fidelity simulators are usually used as closed-loop benchmarking platforms to test and validate autonomous driving modules, including sensor perception, vehicle localization, traffic prediction, path planning, and vehicle control. However, testing and evaluating ITS applications usually involve a large number of AVs along with other road users. Using high-fidelity simulators are highly resource-intensive due to the computational cost of generating sensor signals and rendering the 3D environment at a high frequency (e.g., more than 10 Hz). Especially for network-level applications such as AV-based routing \cite{zhang2018mitigating,guo2021mixed}, there may exist dozens if not hundreds of AVs in the network at the same time, making real-time generation of sensor data and rendering the physical environment impossible, even with high-performance computation platforms.

As another mainstream approach, microscopic traffic simulators only focus on vehicle-level driving behaviors and do not natively support raw sensor data simulation or perception modeling. As illustrated in Fig.~\ref{fig:simulators}b, simplified or aggregated detection surrogates are typically adopted to approximate AV perception results. For example, Li et al. \cite{li2021cooperative} assumed "perfect" detection, which means all vehicles can be detected accurately within the sensor range of AVs. To take a step forward, Li et al. \cite{li2024a} incorporated the average precision (AP) results of a LiDAR detection algorithm on the Waymo Open Dataset (WOD) \cite{WOD} into SUMO simulation. Specifically, vehicles within the AV detection range are randomly dropped (i.e., considered to be unobserved) based on a distance-increasing probability obtained from a LiDAR detection algorithm. Recently, researchers \cite{cao2022analytical} incorporated geometry-based physical visibility to model occlusion using rule-based approaches. As shown in Fig.~\ref{fig:simulators}c, for each AV, surrounding vehicles are sorted by distance, and each vehicle blocks a continuous angular sector in the 2D field of view determined by its bounding-box corners. A vehicle is considered occluded if all its corner angles fall within previously blocked sectors, and otherwise it is detected. However, such 2D, rule-based modeling may overestimate detection occlusions, as it neglects the height dimension and the actual sensing mechanism of range sensors (e.g., the emission and reception of LiDAR beams). As a result, detection outcomes produced by these simplified detection models in Fig.~\ref{fig:simulators}b and Fig.~\ref{fig:simulators}c remain notably different from those obtained by applying object detection algorithms directly to raw sensor data in Fig.~\ref{fig:simulators}a.

\begin{figure}[htbp]
    \centering
    \includegraphics[scale=0.65]{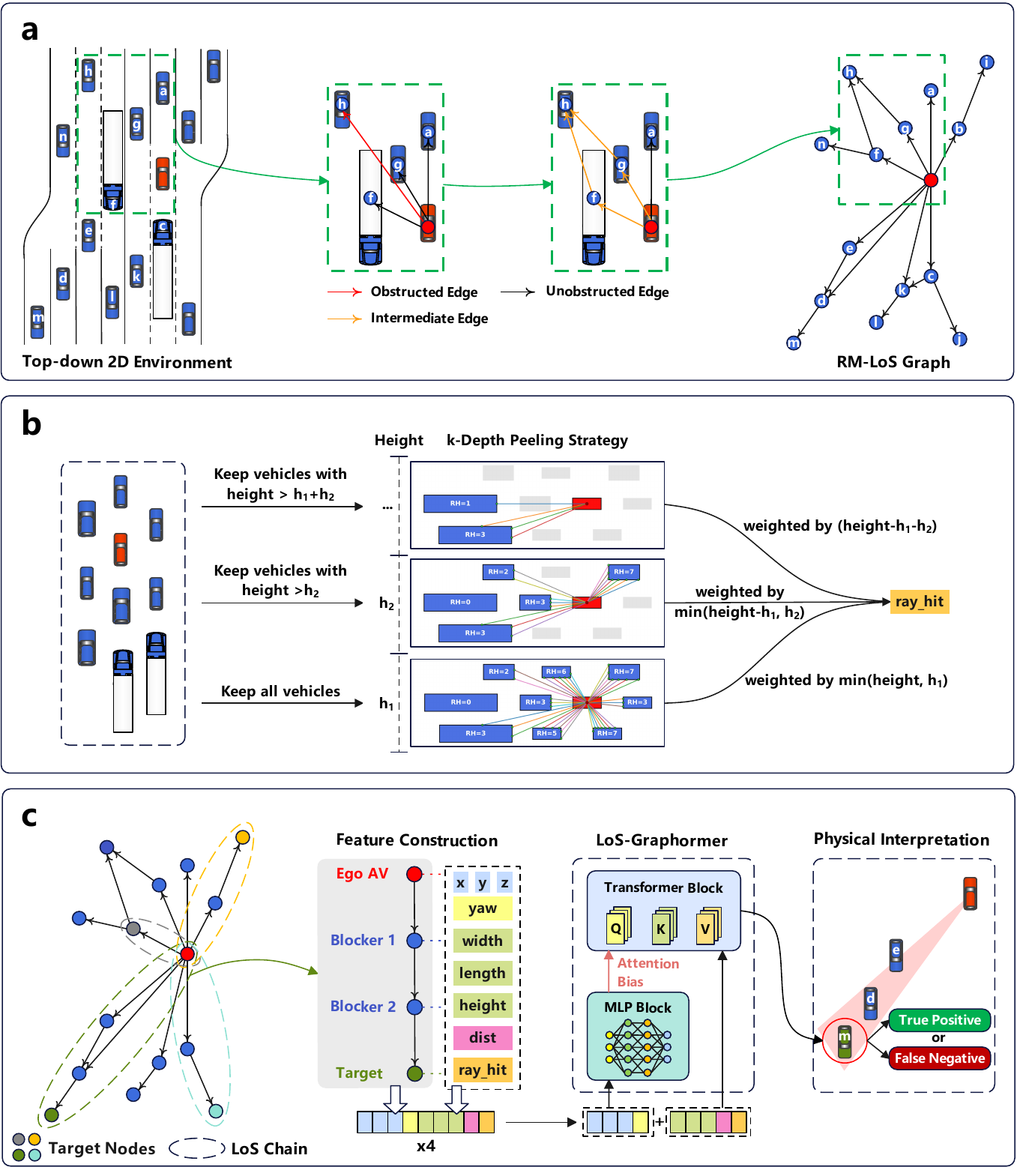}
    \caption{\textbf{The proposed MIDAR framework.}
    \textbf{a.} Construction of the Refined Multi-hop Line-of-Sight (RM-LoS) graph.
    \textbf{b.} Calculation of the ray-hit feature using a Height-aware Azimuthal Ray Casting (HARC) approach.
    \textbf{c.} MIDAR workflow.}
    \label{fig:methodology}
\end{figure}

In summary, while high-fidelity simulators can generate realistic detection results from AVs as in real-world deployment, they face scalability challenges when applied to scenarios involving large networks and multiple AVs. On the other hand, microscopic traffic simulators can efficiently accommodate a large number of AVs, but struggle to approximate realistic detection outputs. The resulting discrepancy between the simplified and realistic detection outputs may further influence the downstream application-level performance.

To reconcile the trade-off between scalability and detection realism, we propose MIDAR, a graph-transformer-based surrogate LiDAR detection model that mimics realistic LiDAR perception using only vehicle-level features available in microscopic traffic simulators. MIDAR constructs a Refined Multi-hop Line-of-Sight (RM-LoS) graph to explicitly encode inter-vehicle relationships such as occlusion, a significant factor in LiDAR detection. As illustrated in Fig.~\ref{fig:methodology}a, for each target vehicle, an LoS chain is generated by recursively tracing its occluding ancestors, from the target vehicle through intermediate blockers to the ego AV. This ordered chain explicitly captures the set of vehicles that may obstruct the ego AV’s line of sight to the target. These LoS chains serve as the fundamental units for the transformer’s attention mechanism. By restricting attention to vehicles along each line-of-sight corridor, MIDAR effectively filters out irrelevant interactions and redundant context, substantially reducing computational complexity while preserving occlusion realism, as illustrated in Fig.~\ref{fig:methodology}c. 

While most of the node features are readily available from microscopic traffic simulators to construct the RM-LoS graph, including vehicle location (x,y,z), dimension (width, length, height), heading (yaw), and their direct distances to the ego AV (dist), MIDAR also introduces a ray-hit feature to include a physical prior on LiDAR visibility using a Height-aware Azimuthal Ray Casting (HARC) approach, as shown in Fig.~\ref{fig:methodology}b. Inspired by the LiDAR detection principle, where object detectability mainly depends on the LiDAR point coverage, this feature serves as a lightweight surrogate by quantifying how many azimuth rays from the ego AV's LiDAR can reach to a target vehicle. To account for vehicle height differences and LiDAR beam tilt, the feature is further extended via a height-aware k-depth peeling strategy, in which the ray casting is performed at multiple height slices and aggregated into a weighted average.

As shown in Fig.~\ref{fig:methodology}c, the ray-hit feature is concatenated to other features and then fed into a LoS-Graphormer model proposed in this study. Specifically, vehicle positions and headings are used to compute pairwise geometric relations, which are passed through a Multi-Layer Perceptron (MLP) to generate an attention bias. This bias injects physically meaningful priors, such as the higher influence of closer vehicles, into the Transformer, improving training stability and inference robustness of MIDAR. All remaining node features are processed through the standard Transformer blocks. For each LoS chain, MIDAR predicts whether the target vehicle (i.e., the last token in the sequence) corresponds to a true positive (TP) or a false negative (FN) detection, thereby mimicking the behavior of a specified object detection algorithm. False positives (FPs) are not considered in MIDAR, as they are comparatively rarer than FNs and require additional modeling assumptions beyond the scope of this work.

In this study, MIDAR is trained and evaluated to mimic the CenterPoint \cite{yin2021center}, a mainstream LiDAR object detection model, using both simulated and real-world point cloud data. To demonstrate its accuracy and practical value, we further evaluate MIDAR through two cooperative-perception-based (CP-based) ITS applications: CP-based adaptive traffic signal control and vehicle trajectory reconstruction. The results show that MIDAR produces substantially different application-level performance metrics than simplified detection models, while generating very close performance metrics as the real LiDAR detection model. These case studies also demonstrate that MIDAR can be seamlessly integrated into not only microscopic traffic simulators such as SUMO, but also other data sources that provide vehicle-level features (e.g., trajectory datasets), to generate detection outputs that closely reflect real-world sensing behavior. 

Our major contributions are outlined below:
\begin{enumerate}
    \item	The proposed MIDAR model is one of the first works that attempt to close the gap between scalability and fidelity (i.e., detection realism) for AV perception simulation.
    \item   MIDAR is lightweight and efficient, and can be easily incorporated into microscopic traffic simulators and other data sources with vehicle-level features. It achieves substantially higher computational efficiency while requiring orders-of-magnitude fewer GPU and CPU resources than game-engine-based simulators such as CARLA.
    \item	MIDAR shows high accuracy in approximating the detection results of a mainstream LiDAR detection model, validated on both real-world and simulated point cloud data. 
    \item	Experiments with two CP-based ITS applications validate the necessity and practical value of the proposed surrogate sensor model, by showing incorporating MIDAR can generate more accurate application-level performance metrics. 
    \item	Although designed for AV LiDAR simulation, the proposed sensor surrogate modeling framework can be generalized to other sensor modalities (e.g., cameras and multi-sensor fusion) and applied to other cyber-physical system (CPS) simulations such as robotics.
\end{enumerate}

\section{Results}

\subsection{Datasets and Target Detection Model}

\subsubsection{NuScenes Dataset}
The nuScenes dataset \cite{caesar2020nuscenes} is a widely used large-scale public dataset for autonomous driving (AD) evaluation. It comprises 1,000 driving scenes collected in Boston and Singapore, two cities that are known for their dense traffic and highly challenging driving situations. The dataset contains a diverse set of traffic scenarios across different locations, weather conditions, vehicle types, and both left- and right-hand traffic. The LiDAR data in nuScenes are collected using a Velodyne HDL-32E spinning LiDAR. Its main specifications are summarized in Table~\ref{tab:hdl32e} of Appendix~\ref{appendix:lidarspec}.

\subsubsection{CARLA Dataset}
\begin{figure}[!htbp]
    \centering
    \includegraphics[scale=0.7]{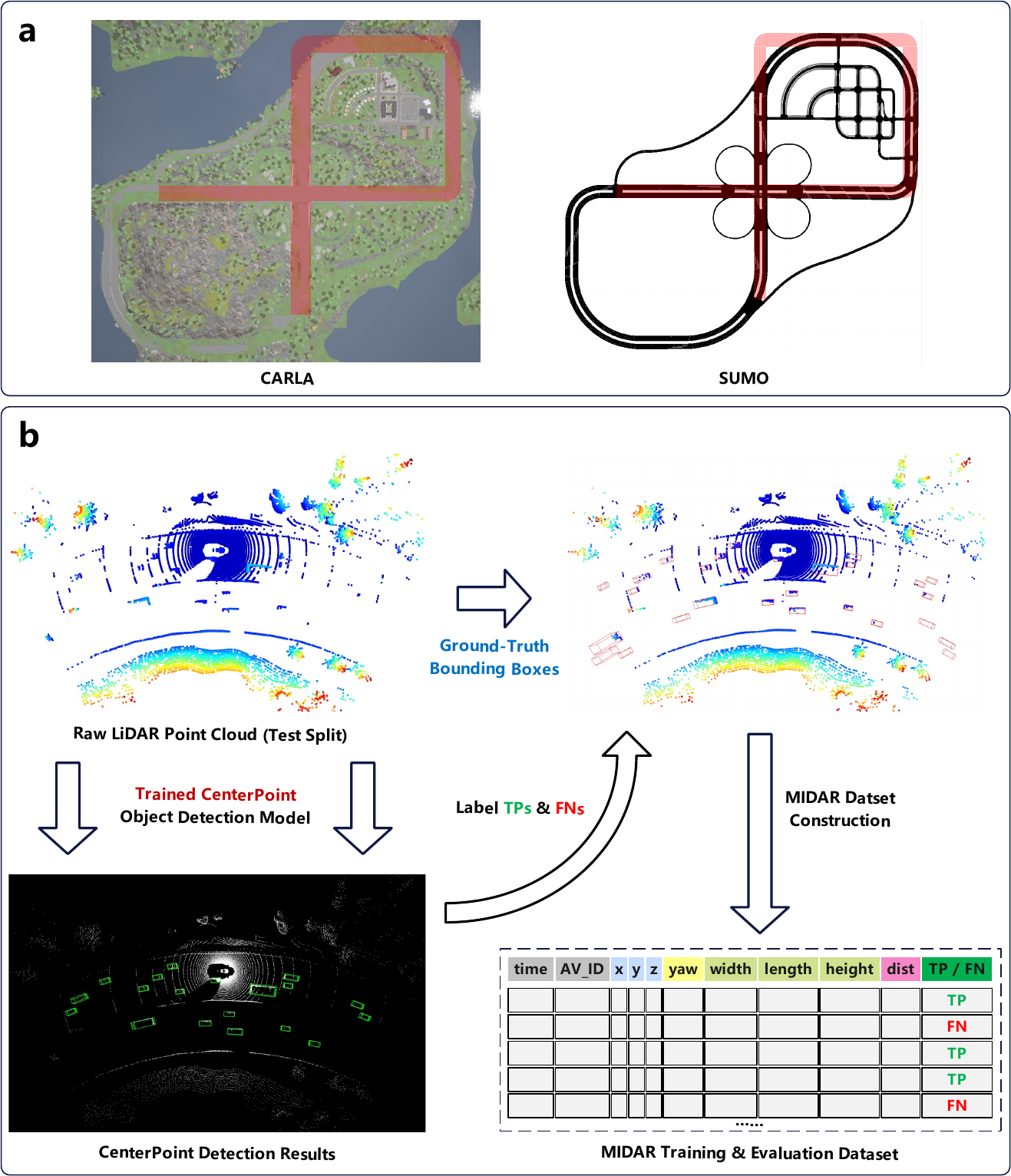}
    \caption{\textbf{Data Preparation for MIDAR Evaluation.}
    \textbf{a.} Data collection area in CARLA-SUMO co-simulation.
    \textbf{b.} Construction of MIDAR training and evaluation dataset.}
    \label{fig:datapreparation}
\end{figure}

Beside the real-world LiDAR dataset, we build a simulated LiDAR dataset using the CARLA–SUMO co-simulation platform \cite{CARLA_SUMO_OAC} to test the performance of MIDAR. Background traffic is generated by SUMO, while CARLA provides high-fidelity LiDAR point cloud data with ground-truth 3D bounding boxes. As illustrated in Fig.~\ref{fig:datapreparation}a, data are collected in a four-lane highway environment in CARLA Town 4 with varying traffic demand levels (2000–12000 veh/h) and include straight, curved, and tunnel segments. The traffic in the simulation is modeled as a heterogeneous fleet comprising 70\% passenger vehicles (sedans and SUVs), 25\% trucks, and 5\% buses. Following the nuScenes data collection paradigm, the dataset is organized into independent scenes, each consisting of continuous LiDAR sweeps captured at a fixed frequency over a short temporal window. Specifically, 300 scenes are generated by combining different traffic demands and vehicles, with 10 Hz frequency recorded for 15 seconds per scene, resulting in a total of 45,000 frames. The LiDAR sensor specifications are summarized in Table~\ref{tab:carla_LiDAR} of Appendix~\ref{appendix:lidarspec}.

\subsubsection{CenterPoint}
CenterPoint \cite{yin2021center}, a widely adopted and well-established LiDAR-based 3D object detection model, is used to generate reference detection results on both nuScenes and CARLA datasets, which MIDAR is trained to mimic. CenterPoint employs a 3D backbone to extract bird-eye-view (BEV) features from LiDAR point clouds, followed by a 2D convolutional neural network (CNN) head to detect object centers and regress 3D bounding boxes. It further refines box predictions using point features extracted at the centers of each box face, passed through a multilayer perceptron (MLP) for Intersection-over-Union (IOU)-guided confidence scoring and box refinement.

The nuScenes and CARLA datasets are first used to train the CenterPoint detectors and generate reference detection results. Specifically, the nuScenes dataset is split into 70\% for training, 15\% for validation, and 15\% for testing, while the CARLA dataset is directly split into 50\% for training, 50\% for testing. We implement the CenterPoint with OpenPCDet \cite{openpcdet}. The detection range is set to 54 meters for nuScenes and 80 meters for the CARLA dataset. The performance of CenterPoint is evaluated on the testing split, with detection performance presented in Table~\ref{tab:centerpoint_perf_nuscenes} and Table~\ref{tab:centerpoint_perf_carla}. In the nuScenes dataset, the performance is reported using the official nuScenes metrics \cite{caesar2020nuscenes}, which compute AP based on center-distance matching, along with complementary localization and attribute error measures. In the CARLA dataset, all vehicle types are treated as a single “car” class to better align with MIDAR's objective (i.e., identifying TPs and FNs independent of vehicle classes) and evaluated using IoU-based AP \cite{geiger2013vision}. It can be seen that CenterPoint achieves strong performance on both the nuScenes and CARLA datasets under their respective evaluation protocols. The results confirm that CenterPoint provides reliable and consistent detection outputs that serve as an appropriate target detection model for training and evaluating MIDAR.

\begin{table}[!htbp]
\centering
\caption{Performance of CenterPoint on the nuScenes dataset.}
\label{tab:centerpoint_perf_nuscenes}
\begin{tabular}{lcccccc}
\hline
\textbf{Class} & \textbf{AP} & \textbf{ATE} & \textbf{ASE} & \textbf{AOE} & \textbf{AVE} & \textbf{AAE} \\
\hline
Car & 0.850 & 0.179 & 0.155 & 0.149 & 0.206 & 0.183 \\
Truck & 0.567 & 0.315 & 0.181 & 0.139 & 0.188 & 0.216 \\
Bus & 0.703 & 0.330 & 0.185 & 0.100 & 0.343 & 0.263 \\
Trailer & 0.382 & 0.516 & 0.196 & 0.595 & 0.166 & 0.179 \\
Construction Vehicle & 0.181 & 0.731 & 0.423 & 0.899 & 0.118 & 0.305 \\
\hline
\end{tabular}
\end{table}

\begin{table}[!htbp]
\centering
\caption{Performance of CenterPoint on the CARLA dataset.}
\label{tab:centerpoint_perf_carla}
\begin{tabular}{lcc}
\hline
\textbf{Class} & \textbf{AP@IoU=0.7} & \textbf{AP@IoU=0.5} \\
\hline
Car  & 70.57 & 72.63 \\
\hline
\end{tabular}
\end{table}

\subsection{MIDAR Dataset Preparation}
As described earlier, MIDAR approximates LiDAR detection results by classifying each ground-truth (GT) vehicle within the sensing range as either a true positive (TP) or a false negative (FN). The detection results of the CenterPoint model on the test splits of both datasets are used as labels for MIDAR's training and evaluation. 
The predicted bounding boxes from the CenterPoint model are matched against GT annotations to assign TP or FN labels to each surrounding vehicle. These labels, together with the corresponding vehicle-level features, constitute the final dataset used to train and evaluate MIDAR, as illustrated in Fig.~\ref{fig:datapreparation}b.

For the nuScenes dataset, vehicles and objects outside drivable areas (e.g., roadways) are removed from both predictions and ground-truth annotations using the nuScenes API and map information. This step pre-processes the detection results to better align with traffic simulation environments and simplifies MIDAR’s learning task. No additional preprocessing is applied to the CARLA dataset, since all vehicles are generated within the roadway boundaries in CARLA simulation.

The two datasets follow the same procedure to label TPs and FNs on the GT vehicles.Specifically, the Hungarian algorithm \cite{kuhn1955hungarian} is applied to maximize the total IoU between the GT bounding boxes and the predicted bounding boxes at the same timestamp. Before matching, predicted bounding boxes are filtered to retain only those with confidence scores above a threshold (0.3 for both datasets). After matching, the GT–prediction pairs are kept only if their IoU exceeds 0.5. Matched GT bounding boxes are labeled as TPs, while the unmatched GT bounding boxes are labeled as FNs. The remaining unmatched predicted bounding boxes are labeled as FPs. The detailed procedure is summarized in Algorithm~\ref{alg:iou_hungarian_labeling}.

On average, the CARLA dataset contains 13.31 TPs, 4.48 FNs, and 1.49 FPs per frame, whereas the nuScenes dataset, which features more complex traffic scenes, contains 6.81 TPs, 3.30 FNs, and 2.65 FPs. These statistics support our assumption that FPs occur less frequently than FNs.

\subsection{Training and Evaluation of MIDAR}
The testing splits of the nuScenes and CARLA datasets are used to construct two MIDAR datasets, each of which is further split into 70\% for training, 15\% for validation, and 15\% for testing. 

The LoS-Graphormer, which serves as the backbone of MIDAR, uses a two-layer MLP for attention bias and a three-layer Transformer encoder. Each Transformer layer uses a hidden dimension of 128, four attention heads, and a position-wise feedforward network with dimension 256. The ray-hit feature is calculated using 720 rays per height slice and the height-aware peeling is applied at two thresholds (1.8m and 3.6m), which roughly correspond to typical passenger-vehicle and truck heights.

The configurations for LoS-Graphormer and all the baselines below are the same for the two datasets.

To evaluate the effectiveness of the proposed architecture, we compare MIDAR with three baselines:
\begin{enumerate}
    \item \textbf{MLP}: a three-layer multilayer perceptron with a hidden dimension of 256 that treats each vehicle independently and ignores the geometric relational structure.
    \item \textbf{GCN}: a graph convolutional network with three convolution layers and a hidden dimension of 256, applied to the same LoS chains with local message passing.
    \item \textbf{Vanilla Transformer}: a Transformer encoder with the same configuration as the LoS-Graphormer but without attention bias.
\end{enumerate}

As introduced earlier, in the LoS-Graphormer, vehicle positions (x,y,z) and headings are used exclusively to construct the attention bias and are not provided as direct inputs to the prediction module. To ensure a fair comparison, all baseline models therefore use the same five input features (vehicle height, width, length, distance to the ego AV, and ray-hit) for prediction. All models are trained using the AdamW optimizer \cite{loshchilov2017decoupled} with a learning rate of $2\times10^{-4}$ and a weight decay of $10^{-4}$. To address the class imbalance between TP and FN, we employ a focal loss with a focusing parameter $\gamma = 2.0$ and class weights $\boldsymbol{\alpha}$ computed from the empirical class distribution. The training process uses 500 warm-up steps and early stopping with a patience of 10 epochs, with a maximum of 200 training epochs.

We primarily evaluate model performance using the area under the ROC curve (AUC) for TP/FN classification, and additionally report precision, recall, and F1 score to provide a comprehensive assessment. All metrics other than AUC are computed on the testing set using the decision threshold that maximizes the F1 score, which balances precision and recall. The results on the CARLA and nuScenes datasets are respectively demonstrated in Table~\ref{tab:carla_model_comparison} and Table~\ref{tab:nuscenes_model_comparison}.

From the results, we can see that, in general, the proposed LoS-Graphormer model (i.e., MIDAR) consistently outperforms all baseline models across both datasets, demonstrating its effectiveness in mimicking the CenterPoint LiDAR detection behaviors. On the CARLA dataset, which is generated from a more uniform and controllable simulation environment, the LoS-Graphormer achieves an AUC close to 0.94, indicating strong discrimination between TPs and FNs. On the nuScenes dataset, which contains more complex and noisy real-world scenes, performance decreases moderately to an AUC of 0.86. Furthermore, we conduct an ablation test regarding the impact of the ray-hit feature, as shown in the last row of the two tables. It can be seen that removing the ray-hit feature in the CARLA dataset significantly downgrades the MIDAR performance with decreased metrics in all aspects. However, in the nuScenes dataset, the difference becomes less prominent. This suggests that more diverse vehicle composition, environmental noise, and sensing uncertainties in real-world driving environments may influence the function of the ray-hit feature in the LoS-Graphormer model. However, the application-level evaluations show how the physical priors introduced by ray-hit can provide tangible benefits.

\begin{table}[t]
\centering
\caption{Performance comparison of models on the CARLA dataset.}
\label{tab:carla_model_comparison}
\begin{tabular}{lcccc}
\hline
\textbf{Model} & \textbf{AUC} & \textbf{F1} & \textbf{Precision} & \textbf{Recall} \\
\hline
MLP & 0.8629 & 0.5666 & 0.5135 & 0.7410 \\
GCN & 0.8352 & 0.5725 & 0.4815 & 0.7059 \\
Vanilla Transformer & 0.9135 & 0.6974 & 0.6782 & 0.7178 \\
LoS-Graphormer (w/ ray-hit) & \textbf{0.9385} & \textbf{0.7576} & \textbf{0.7665} & \textbf{0.7490} \\
LoS-Graphormer (w/o ray-hit) & 0.8982 & 0.6625 & 0.6137 & 0.7199 \\
\hline
\end{tabular}
\end{table}

\begin{table}[t]
\centering
\caption{Performance comparison of models on the nuScenes dataset.}
\label{tab:nuscenes_model_comparison}
\begin{tabular}{lcccc}
\hline
\textbf{Model} & \textbf{AUC} & \textbf{F1} & \textbf{Precision} & \textbf{Recall} \\
\hline
MLP & 0.8285 & 0.6765 & 0.5823 & 0.8071 \\
GCN & 0.7967 & 0.6438 & 0.5495 & 0.7773 \\
Vanilla Transformer & 0.8378 & 0.6858 & 0.5879 & 0.8230 \\
LoS-Graphormer (w/ ray-hit) & 0.8647 & 0.7079 & \textbf{0.6380} & 0.7951 \\
LoS-Graphormer (w/o ray-hit) & \textbf{0.8665} & \textbf{0.7153} & 0.6246 & \textbf{0.8366} \\
\hline
\end{tabular}
\end{table}

\subsection{Application-level Evaluation}
Since the main motivation for MIDAR is to enable realistic detection modeling for ITS applications, application-level evaluation is essential to justify the motivation and validate its performance. In this section, we integrate MIDAR with two representative ITS applications and compare its performance against a real LiDAR detection model (i.e., CenterPoint) and two commonly used simplified baseline detection models using application-specific performance metrics. We first introduce the two baseline models that are typically adopted in previous studies as below:

\begin{enumerate}
\item \textbf{Perfect Detection}: all surrounding vehicles within the ego-vehicle's LiDAR detection range are assumed to be observed without errors. The Perfect Detection model is equivalent to 100\% TPs and 0\% FNs.
\item \textbf{Random Dropout}: Vehicles within the LiDAR detection range are randomly dropped (i.e., labeled as false negatives), with the dropout probability increasing as the distance to the ego vehicle increases. The detection range is partitioned into distance intervals, and a dropout probability is assigned to each interval. In this study, these probabilities are derived from the False Negative Rates (FNRs) of the CenterPoint detector evaluated on the raw LiDAR point cloud datasets, ensuring that the distance-dependent detection reliability is grounded in empirical detector performance rather than heuristic assumptions.
\end{enumerate}

\begin{figure}[!htbp]
    \centering
    \includegraphics[scale=0.24]{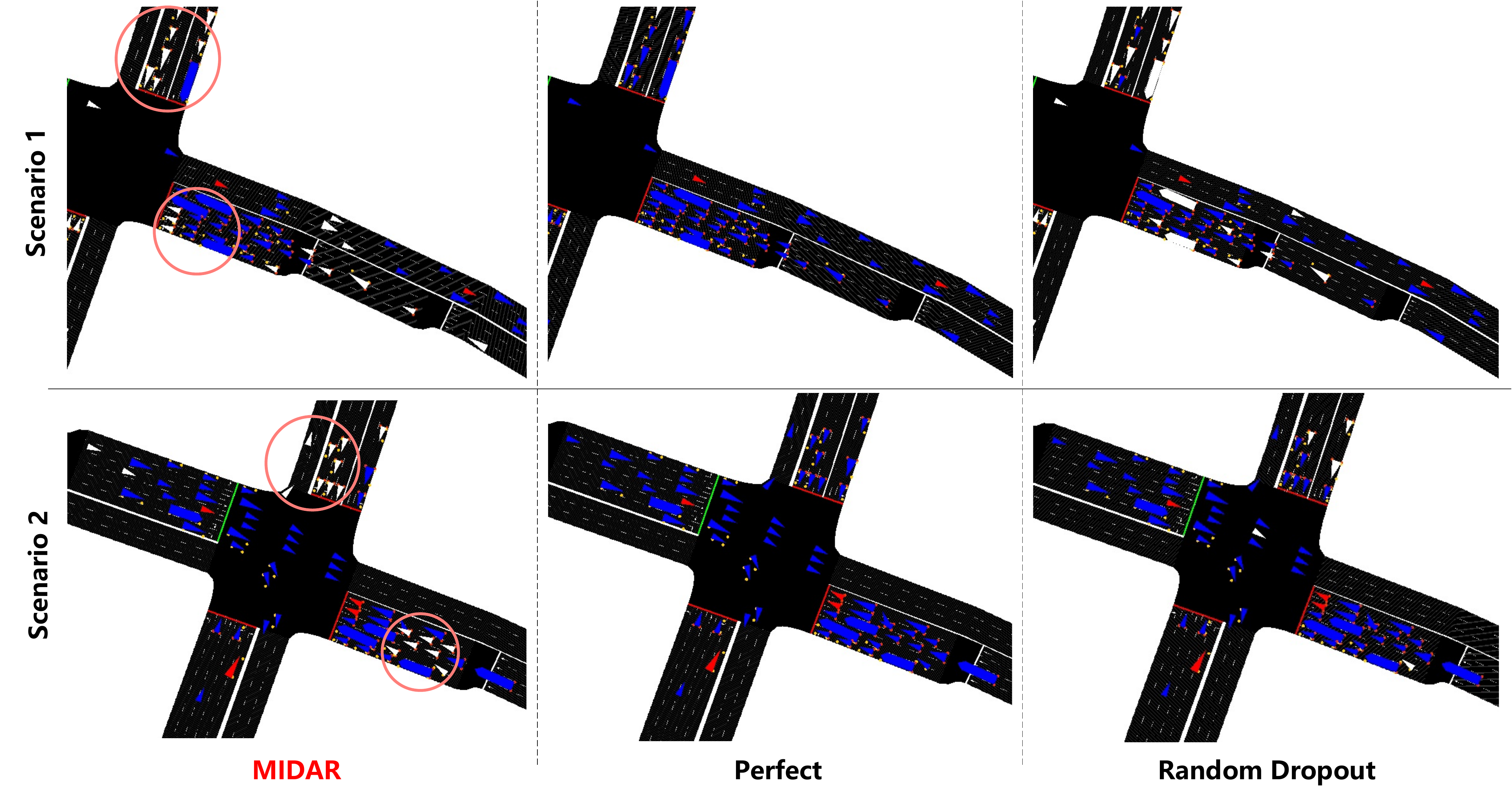}
    \caption{\textbf{Comparison between MIDAR and two baseine models}}
    \label{fig:detection_comaprison}
\end{figure}

Fig.~\ref{fig:detection_comaprison} provides an example comparison of MIDAR and two baseline detection models under two different scenarios in a microscopic traffic simulator. In the figure, red vehicles denote AVs, blue vehicles indicate those detected by the AVs, and the remaining vehicles represent undetected background traffic. Overall, MIDAR exhibits the most plausible detection patterns, particularly in the regions highlighted by the pink circles. The perfect detection model unrealistically detects nearly all surrounding vehicles regardless of occlusions, whereas the random dropout model produces unrealistic patterns, such as failing to detect nearby vehicles while detecting farther or occluded ones.

\subsubsection{Application 1: Traffic Signal Control}

\paragraph{Application Background}:
Traffic signal control (TSC) plays a critical role in managing and optimizing the flow of vehicles at intersections. Traditional approaches rely on fixed-location detectors (e.g., loop detectors), which provide limited spatial coverage and may fail to capture the full dynamics of traffic flow, thereby reducing control effectiveness. Recent studies such as \cite{li2024a} have demonstrated that CP-based TSC, which fuses observations from multiple AVs near intersections, can substantially improve the control performance even under very low market penetration rates (MPRs).

\begin{figure}[!htbp]
    \centering
    \includegraphics[scale=0.87]{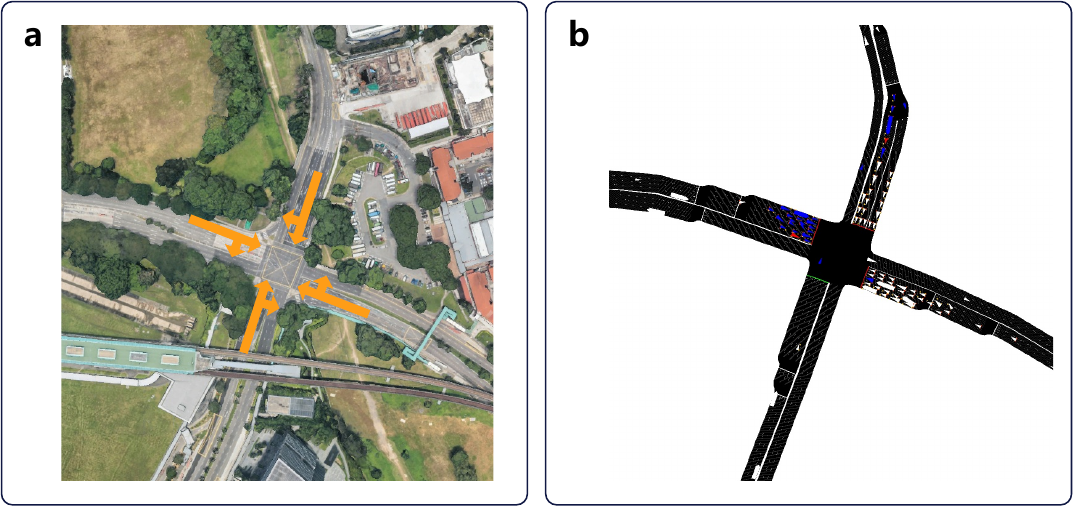}
    \caption{\textbf{Study area of CP-based adaptive TSC.}
    \textbf{a.} A real-world intersection at Commonwealth Avenue and North Buona Vista Road in Singapore.
    \textbf{b.} The intersection in SUMO.}
    \label{fig:application_tsc}
\end{figure}

\paragraph{Application Setup}
A real-world intersection at Commonwealth Avenue and North Buona Vista Road in Singapore is selected as the study area for the traffic signal control application, as illustrated in Fig.~\ref{fig:application_tsc}. The main reason for selection of this particular intersection is that the nuScenes dataset was collected at the same location. As a result, we use the MIDAR model trained on the nuScenes dataset as well as the baseline detection models to conduct the experiments. We select SUMO \cite{Lope2018sumo} as the microscopic traffic simulator. The corresponding SUMO network of the intersection is also shown in Fig.~\ref{fig:application_tsc}a. The integration of the MIDAR model and SUMO simulation environment is established through SUMO's Traffic Control Interface (TraCI) \cite{Lope2018sumo}. For each AV at each simulation time step, one selected LiDAR detection model (Perfect Detection, Random Dropout, or MIDAR) is applied. The maximum detection ranges for the three models are consistently 54 m. As mentioned before, the set of dropout probabilities for Random Dropout within different distance intervals are aggregated based on the FNRs of CenterPoint detection results on the nuScenes dataset within different distance intervals: (1) 0-10m: 3.1\%; (2) 10-20m: 6.9\%; (3) 20-30m: 22.6\%; (4) 30-40m: 43.4\%; (5) 40-50m: 64.7\%; and (6) 50-54m: 79.3\%. After obtaining the detection results of each AV (i.e., determination of observed and unobserved vehicles), a simple fusion algorithm is applied to remove redundant vehicles and generate a collective CP observation. Fig.~\ref{fig:application_tsc}b shows a snapshot of the MIDAR detection results where the red vehicles are AVs, blue vehicles are detected surrounding vehicles, and white vehicles are unobserved vehicles.

To further reflect real world traffic scenarios, traffic volumes at the intersection are obtained and calibrated through the Singapore Land Transport Authority’s Data Mall API \cite{LTAdatamall}. Additionally, the vehicle composition in SUMO is configured to match the vehicle class distribution observed in the nuScenes dataset.
A split 4-phase signal timing plan is used, represented by orange arrows in Fig.~\ref{fig:application_tsc}a, same as in real-world operations. Each phase corresponds to movements from the Eastbound, Westbound, Southbound, and Northbound directions, with protected right turns (for left-hand traffic). A 5-second minimum green time and a 40-second maximum green time are adopted. A 4-second yellow and a 1-second all-red clearance interval are enforced between each phase transition.

The I-SIG algorithm \cite{feng2015a} is selected as the adaptive signal control model. The I-SIG algorithm utilizes dynamic programming (DP) to optimize signal phase durations and sequences with the objective of minimizing total delay. To solve the DP problem, an arrival table is constructed based on the collected traffic data (i.e., red and blue vehicles in Fig.~\ref{fig:application_tsc}b) and serves as the input, which comprises predictions of future arrival flows for each phase at each timestamp. In this study, the arrival table can be directly constructed using observed vehicle information from the CP environment.
Each simulation run lasts for 3600s with a 0.1s resolution. The first 100 s are considered as the warm-up stage when the signal is controlled by a fixed-time plan. The vehicle arrival follows a Poisson process generated by SUMO. We set the MPR of AVs to be 3\% which is similar as in \cite{li2024a}. This MPR represents an early stage of AV deployment. 


\paragraph{Performance Comparison}
We evaluate the performance of the traffic signal control application using average vehicle delay as the primary metric. After running 10 rounds of simulation with different random seeds, Table~\ref{tab:TSC_perf_comparison} summarizes the average vehicle delay of I-SIG under different LiDAR detection models. The average delays for Perfect Detection, Random Dropout, MIDAR with ray-hit, and MIDAR without ray-hit are 58.12 s, 63.57 s, 76.79 s, and 84.09 s, respectively. In most scenarios, both MIDAR models lead to substantially higher delays than the two simplified baselines, indicating that incorporating realistic LiDAR detection effects can significantly degrade the signal control performance compared with idealized perception assumptions. Paired t-tests further confirm that the differences in average delay between MIDAR and Perfect Detection (p=0.0003) and between MIDAR and Random Dropout (p=0.0030) are statistically significant at the 5\% level. Moreover, removing the ray-hit feature further increases both the mean delay and its variance across scenarios, suggesting that neglecting physically grounded visibility information may lead to optimistic bias and overestimation of control performance. 

We can also interpret these results from the traffic signal control model's perspective. Under the perfect detection model, all vehicles within the AV's detection range can be observed, which provides the richest information to the control model and results in the best performance. Under both random dropout and MIDAR models, although the average number of observed vehicles is similar since the distance-based dropout rates are calibrated based on the real LiDAR detection model. However, the missing data patterns differ. Under the random dropout model, undetected vehicles are randomly distributed across both time and space domains, whereas under the MIDAR models, missing detections tend to be structured and can persist over time due to realistic sensing effects, as shown in Figure \ref{fig:detection_comaprison}. For example, if an AV is following a truck or a bus, occlusion may block its downstream vehicles, creating a consistent "empty" area without any detection. This consistent gap will mislead the traffic prediction algorithm to underestimate total vehicle delay and/or queue length and allocate insufficient green time, which in turn increases actual delay. On the other hand, the prediction error caused by random dropout is less significant because the missing observations are unbiased and intermittent. Some vehicles remain visible within the same region and across successive time steps, allowing the prediction algorithm to recover the true traffic state more accurately.

\begin{table}[t]
\centering
\caption{Average vehicle delay in milliseconds for adaptive TSC under different detection models.}
\label{tab:TSC_perf_comparison}
\begin{tabular}{lcccc}
\hline
\textbf{Run} & \textbf{Perfect} & \makecell[l]{\textbf{Random}\\\textbf{Dropout}} & \makecell[l]{\textbf{MIDAR}\\\textbf{(w/ ray-hit)}} & \makecell[l]{\textbf{MIDAR}\\\textbf{(w/o ray-hit)}} \\
\hline
1  & 51.26 & 55.76 & 81.33 & 110.49 \\
2  & 62.01 & 72.59 & 87.50 & 81.91 \\
3  & 61.24 & 72.13 & 57.03 & 89.04 \\
4  & 57.32 & 65.71 & 73.56 & 68.44 \\
5  & 58.27 & 63.52 & 81.25 & 74.24 \\
6  & 53.37 & 57.53 & 71.25 & 95.06 \\
7  & 56.06 & 67.76 & 79.04 & 96.94 \\
8  & 58.44 & 58.36 & 64.43 & 77.95 \\
9  & 65.41 & 61.79 & 93.08 & 79.24 \\
10 & 57.81 & 60.59 & 79.43 & 67.55 \\
avg $\pm$ std & 
$58.12 \pm 4.12$ & 
$63.57 \pm 5.90$ & 
$76.79 \pm 10.63$ & 
$84.09 \pm 13.70$ \\
\hline
\end{tabular}
\end{table}



\subsubsection{Application 2: CP-based Vehicle Trajectory Reconstruction}

\paragraph{Application Background}
Vehicle trajectories are essential for various traffic applications. However, collecting complete vehicle trajectory data using multiple continuously located video cameras or unmanned aerial vehicles, requires high costs. An alternative approach is to reconstruct full trajectories from partial observations, which offers a more cost-effective solution. Existing studies assume that partial observations are collected either in a connected vehicle (CV) environment \cite{Chen01012021,wei2021a} or in a CP environment \cite{zhang2024vehicle,chen2024macro}. For studies that consider CP environment, usually the Perfect Detection model is applied. Two exceptions are the works from \cite{zhang2024vehicle} and \cite{cao2022analytical}, in which the impact of occlusion is modeled: if a vehicle is entirely in the shadows of other vehicles, it is considered unobservable. However, this is still a simplified model that does not reflect real LiDAR detection patterns.

\begin{figure}[!htbp]
    \centering
    \includegraphics[scale=0.92]{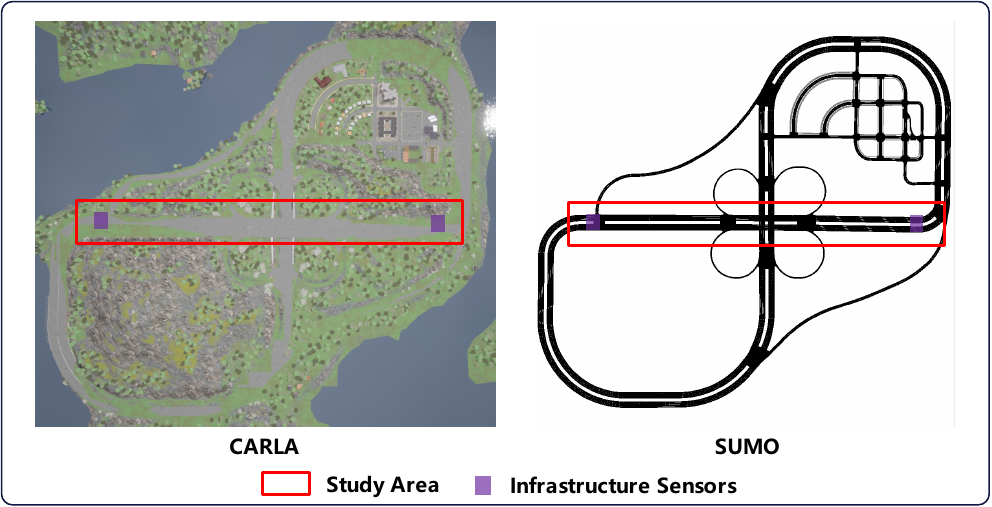}
    \caption{\textbf{Study area of CP-based trajectory reconstruction in CARLA-SUMO co-simulation.}}
    \label{fig:application_traj}
\end{figure}

\paragraph{Application Setup}
Unlike the first application, which directly integrates MIDAR into microscopic traffic simulation, trajectory reconstruction is an offline application. As a result, we evaluate MIDAR using a trajectory dataset with raw LiDAR data collected from the CARLA–SUMO co-simulation platform. The analysis focuses on an 800-meter stretch of the east-bound highway consisting of four lanes in CARLA Town 4, as illustrated in Fig.~\ref{fig:application_traj}. The study area is bounded by infrastructure sensors that define its upstream and downstream limits and provide the initial and final vehicle states required by the trajectory reconstruction model.

An optimization-based model from our recent work \cite{zhu2026cptrajrecon} is employed to reconstruct vehicle trajectories. The model effectively captures lane-change and overtaking behaviors and functions under very low AV MPRs. Specifically, the reconstruction task is formulated as a mixed integer linear programming (MILP) problem with the objective of minimizing the errors between reconstructed and observed vehicle trajectories, which converts the trajectory reconstruction problem into a joint trajectory generation problem. The model incorporates various constraints on decision variables, vehicle dynamics, lane-changing behavior, and safety to ensure realistic and feasible outputs. When implemented, the model simultaneously generates the longitudinal position and lane index of each vehicle at each time step.

To generate CP data, the trajectory dataset collected from the CARLA-SUMO co-simulation platform is preprocessed with an MPR of 2\%, resulting in 40 uniformly sampled AVs. These AVs report both their own trajectories and the detected trajectories of surrounding vehicles. The optimization problem is formulated based on a planning horizon, which is defined as from the first AV entering the study area until the next AV leaving the study area. With the 40 sampled AVs, a total of 39 planning horizons are generated. Because raw LiDAR data are available for the sampled AVs, MIDAR can be directly compared with the CenterPoint detection model training on the CARLA dataset, as well as the two simplified baselines. The probabilities of Random Dropout in this case are aggregated based on the FNRs of CenterPoint detection results on the CARLA dataset: (1) 0-10m: 1.0\%; (2) 10-20m: 5.68\%; (3) 20-30m: 11.4\%; (4) 30-40m: 11.5\%; (5) 40-50m: 18.5\%; (6) 50-60m: 27.1\%; (7) 60-70m: 36.5\%; (8) 70-80m: 48.7\%.

\begin{table}[t]
\centering
\caption{Trajectory reconstruction performance under different detection models.}
\label{tab:traj_recon_perf}
\begin{tabular}{lcccccc}
\hline
\textbf{Detection Model} & \textbf{Recon. Rate} & $\mathbf{MAE}_x$ & $\mathbf{MAPE}_x$ & $\mathbf{RMSE}_x$ & $\mathbf{MAE}_k$ & $\mathbf{MAE}_{lc}$ \\
\hline
\rowcolor{lightred}
Real LiDAR Detection & 48.59\% & 1.525 & 0.721 & 3.564 & 0.0150 & 1.620 \\
MIDAR (w/ ray-hit) & \textbf{46.67\%} & \textbf{1.522} & 0.727 & \textbf{3.522} & \textbf{0.0148} & \textbf{1.628} \\
MIDAR (w/o ray-hit) & 45.61\% & 1.498 & \textbf{0.718} & 3.510 & 0.0154 & 1.689 \\
Random Dropout & 46.01\% & 1.372 & 0.667 & 3.373 & 0.0123 & 1.327 \\
Perfect Detection & 38.42\% & 1.306 & 0.636 & 3.325 & 0.0127 & 1.387 \\
\hline
\end{tabular}
\end{table}

\paragraph{Performance Comparison}
Table~\ref{tab:traj_recon_perf} compares trajectory reconstruction performance under different detection models. The metrics $MAE_x$, $MAPE_x$, and $RMSE_x$ measure the mean absolute error, mean absolute percentage error, and root mean squared error of reconstructed vehicle longitudinal positions relative to ground truth. $MAE_k$ and $MAE_{lc}$ evaluate the accuracy of reconstructed lane positions and lane-change timing, respectively, while the reconstruction rate denotes the proportion of trajectory points successfully reconstructed. Across nearly all metrics, MIDAR with the ray-hit feature produces results closest to those obtained using true LiDAR detection. Paired $t$-tests further confirm this observation. For $MAE_x$, statistically significant differences at the 5\% level are observed between real LiDAR detection and Perfect Detection ($p = 7.8\times10^{-8}$), as well as between real LiDAR detection and Random Dropout ($p = 8.98\times10^{-4}$). In contrast, no significant difference is found between real LiDAR detection and MIDAR, either with ray-hit ($p = 1.0$) or without ray-hit ($p = 0.4$). These results indicate that MIDAR effectively preserves the application-level performance characteristics of real LiDAR detection. These results demonstrate that MIDAR can effectively approximate the application-level performance of real LiDAR detection. Notably, although Random Dropout matches aggregate detection precision and achieves a reconstruction rate comparable to MIDAR, it fails to reproduce realistic spatial detection patterns, leading to significantly different trajectory reconstruction outcomes. This highlights the limitations of simplified detection models and underscores the necessity of our proposed sensor surrogate model. Finally, while the quantitative gains are modest, incorporating the ray-hit feature introduces additional physical grounding and consistently improves MIDAR’s realism in downstream trajectory reconstruction tasks.

\subsection{Computational-cost Evaluation}

We evaluate the computational cost of applying MIDAR directly within SUMO and compare it against a conventional LiDAR perception pipeline deployed under a CARLA--SUMO co-simulation framework. The experiment is conducted in the same study area as shown in Fig.~\ref{fig:application_traj}, under AV MPRs ranging from 5\% to 15\%. All experiments are performed on a workstation with an Intel Core i9-13900KF CPU, an NVIDIA RTX 4090 GPU (24~GB VRAM), and 64~GB of system memory.

During the simulation, CPU resident set size (RSS) memory usage and GPU video memory (VRAM) usage are logged at every simulation tick for all relevant processes. 
In addition, wall-clock times are recorded for each simulation tick (per-tick time) as well as for the detection associated with each AV (per-AV time). 
The per-tick time captures the end-to-end latency of a simulation step, including both the per-AV time and the overhead introduced by the simulation platforms.

For MIDAR, the per-AV time is defined as the total time required to complete the full MIDAR pipeline, including the RM-LoS graph construction, feature assembly, and LoS-Graphormer inference.
In contrast, for the conventional pipeline, the per-AV time only accounts for the execution of the LiDAR-based detection module (CenterPoint in our implementation), while additional overhead from 3D environment rendering, sensor simulation, and point cloud generation is reflected in the per-tick time.

The system-level computational cost comparison is summarized in Table~\ref{tab:system_cost}. 
Overall, MIDAR exhibits a consistently low and bounded resource footprint across all penetration rates. 
Across all evaluated MPRs, MIDAR requires less than 0.5~GB of GPU memory and maintains a nearly constant per-AV detection time of approximately 6--7~ms. In contrast, the conventional perception pipeline incurs substantially higher computational overhead. Specifically, the pipeline consumes over 21~GB of GPU memory and about 8~GB of CPU memory, driven primarily by LiDAR object detection algorithm and CARLA rendering and sensor simulation. 

Taken together, these results demonstrate that MIDAR achieves substantially higher computational efficiency while requiring orders-of-magnitude fewer GPU and CPU resources. Despite using roughly 44$\times$ less GPU memory and 8$\times$ less CPU memory than the conventional pipeline, MIDAR delivers over 3.5$\times$ improvement in per-AV detection time and 3$\times$ improvement in per-tick time under high penetration rates. 
This efficiency gain highlights the advantage of replacing high-fidelity sensor simulation with a lightweight sensor surrogate detection model when large-scale or long-horizon simulations are required.

A detailed process-level breakdown of CPU and GPU resource usage for both approaches at 10\% MPR is provided in Appendix~\ref{app:resource_breakdown}.

\begin{table}[t]
\centering
\caption{System-level computational cost comparison between MIDAR and a conventional perception pipeline under different AV MPRs.}
\label{tab:system_cost}
\setlength{\tabcolsep}{4pt} 
\begin{tabular}{l c cc cc}
\toprule
Method & MPR &
\multicolumn{2}{c}{Memory (GB)} &
\multicolumn{2}{c}{Latency (ms)} \\
\cmidrule(lr){3-4} \cmidrule(lr){5-6}
 &  & CPU & GPU & Per-AV & Per-Tick \\
\midrule
MIDAR & 5\%  & 1.08 & 0.48 & 6.81 & 40.56 \\
MIDAR & 10\% & 1.08 & 0.48 & 6.90 & 59.12 \\
MIDAR & 15\% & 1.08 & 0.48 & 6.71 & 80.45 \\
\midrule
Conventional & 5\%  & 7.88 & 21.25 & 19.25 & 115.26 \\
Conventional & 10\% & 8.01 & 21.27 & 18.18 & 154.23 \\
Conventional & 15\% & 8.13 & 21.27 & 17.71 & 212.58 \\
\bottomrule
\end{tabular}
\end{table}

\section{Discussion}
This study presents MIDAR, a lightweight surrogate LiDAR detection model that bridges the long-standing gap between scalability and perception realism in simulating ITS applications. By leveraging vehicle-level features and physically informed visibility cues, MIDAR enables realistic perception modeling without the computational burden of full sensor simulation. Extensive evaluations on both simulated and real-world datasets demonstrate that MIDAR closely approximates state-of-the-art LiDAR detection performance while remaining robust across domains. Application-level case studies further show that simplified perception assumptions can lead to substantially biased outcomes, underscoring the necessity of realistic detection modeling in cooperative and infrastructure-assisted mobility systems. Importantly, MIDAR integrates seamlessly with microscopic traffic simulators and trajectory datasets, enabling large-scale and accelerated experimentation with perception-aware ITS and AV applications. While developed for LiDAR detection, the proposed surrogate detection framework is generic and can be extended to other sensing modalities as such camera and radar. Moreover, the same modeling approach can be extended to other cyber physical systems (CPS) other than automotive and transportation systems. For example, robotic warehouse automation, where fleets of mobile robots rely on imperfect sensing for localization, obstacle detection, and coordination, can also apply the MIDAR model with a revised feature list to generate realistic sensing results and support efficient simulation of large warehouse operations.

Although challenging, future research could explore how to model FPs, for example, by introducing candidate nodes into the graph based on empirical experience to represent potential FPs.

\section{Methods}

\subsection{Construction of RM-LoS Graph}
\begin{figure}[!htbp]
    \centering
    \includegraphics[scale=0.57]{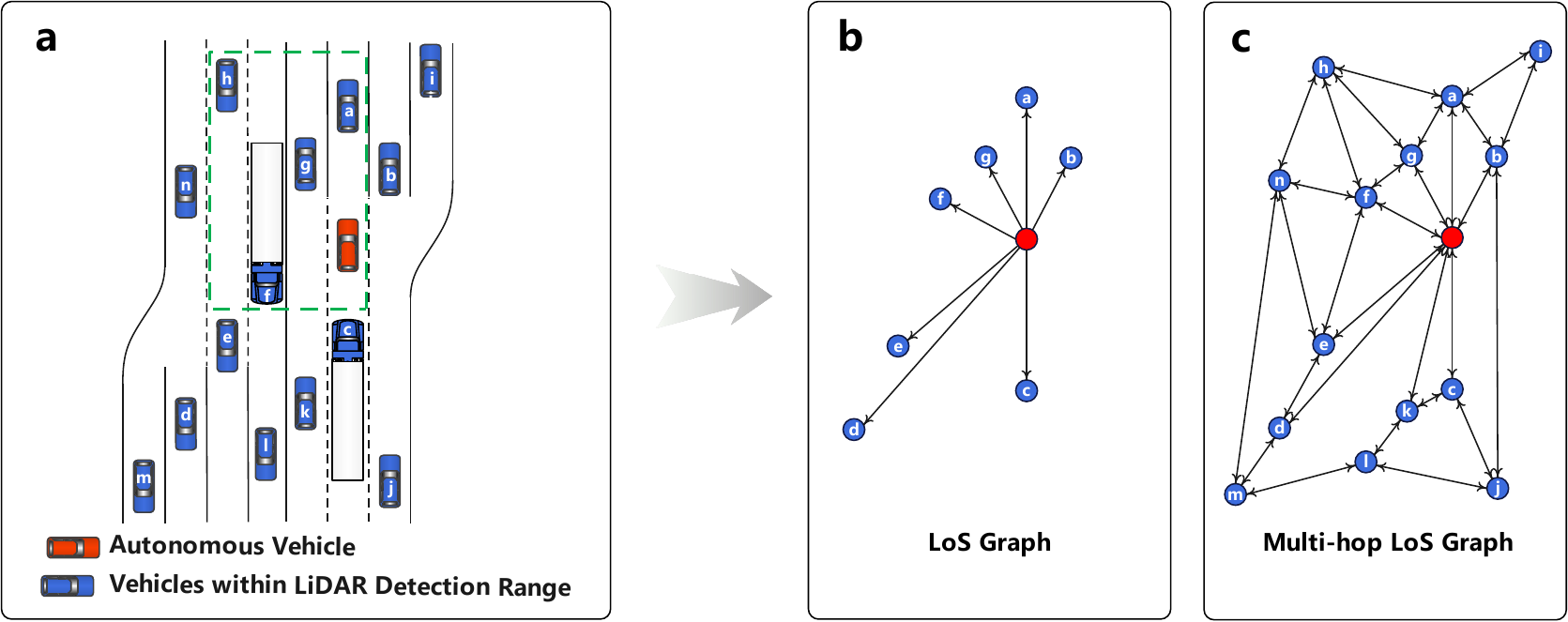}
    \caption{\textbf{Illustration of LoS and Multi-hop LoS graphs}
    \textbf{a.} Bird's eye view of a traffic scene.
    \textbf{b.} A naive Line-of-Sight graph.
    \textbf{c.} A naive multi-hop Line-of-Sight graph.}
    \label{fig:graph_comparison}
\end{figure}

A Refined Multi-hop Line-of-Sight (RM-LoS) graph is constructed to encode occlusion relationships among the ego AV, target vehicles, and potential blockers. The concept of Line-of-Sight (LoS) graph was firstly introduced by \cite{garey1976graph} for testing printed circuit boards for potential short circuits. In the context of autonomous driving, an LoS graph provides a natural abstraction of perception, where surrounding vehicles are connected to the ego AV only if a direct, unobstructed line of sight exists, as illustrated in Fig.~\ref{fig:graph_comparison}b.

While this basic LoS formulation captures direct visibility, it does not fully reflect LiDAR perception, since vehicles that are not directly visible may still be detected through partial occlusions. To address this, the LoS graph can be extended to a multi-hop LoS graph, where vehicles are connected via intermediate vehicles if a sequence of unobstructed LoS segments exists. For example, even if a target vehicle is occluded from the ego AV, it may still be connected through one or more intervening vehicles, as shown in Fig.~\ref{fig:graph_comparison}c. However, naive multi-hop LoS graphs often introduce excessive and non-informative edges, and their bidirectional connections fail to reflect the inherently directional nature of perception, which flows from the ego AV outward.

To overcome these limitations, we propose the RM-LoS graph, which selectively preserves physically meaningful occlusion relationships. As illustrated in Fig.~\ref{fig:methodology}a, the RM-LoS graph is constructed as follows:
\begin{enumerate}
\item	Select one surrounding vehicle within the LiDAR detection range, connect an edge from the center of the AV to the center of the selected vehicle.
\item	If this edge intersects with any other vehicles (i.e., the edge is obstructed), it is segmented into multiple intermediate edges. Specifically, the intermediate edges first originate from the AV to the intersecting vehicles, then from the intersecting vehicles to the target vehicle.
\item	Repeat Steps 1 and 2 for all vehicles.
\end{enumerate}

Using truck “f”, vehicle “g” and vehicle “h”, as an example, the edge connecting vehicle “h” and the AV would intersect with truck “f” and vehicle “g”. Therefore, it’s segmented into 4 intermediate edges (AV to “f”, AV to “g”, “f” to “h”, “g” to “h”), meaning that the detection of vehicle “h” is associated to the features (e.g., locations and sizes) of truck “f” and vehicle “g”. It’s worth noting that 1) the vehicle selection sequence can be arbitrary; and 2) all the constructed edges in RM-LoS are unidirectional, originating from the AV to surrounding vehicles, or from nearer to farther vehicles, to better reflect the occlusion relationship between vehicles.

Under this construction, each target vehicle is associated with a LoS chain, defined as an ordered sequence consisting of the ego AV, the target vehicle, and its occluding ancestors. These LoS chains explicitly capture the vehicles that may influence the visibility of the ego AV to the target and serve as the fundamental units for the LoS-Graphormer, enabling efficient attention over physically relevant occlusion relationships while pruning redundant connections.

\subsection{Feature Selection}

As illustrated in Fig.~\ref{fig:methodology}c, most node features in the RM-LoS graph are readily available from the microscopic traffic simulators, including vehicles’ 3D position in the Cartesian coordinates of the LiDAR sensor frame (i.e., x, y, z), physical dimensions (i.e., width, length, height), heading, and Euclidean distance from the ego AV. Besides those features, to provide a lightweight yet physically informed visibility prior for surrogate LiDAR detection, we construct the ray-hit feature using a Height-Aware Azimuthal Ray Casting (HARC) approach, as shown in Fig.~\ref{fig:methodology}b. The key motivation of incorporating this feature is explained as follows. In a real LiDAR detection pipeline, an object is more likely to be detected when it receives and reflects sufficient LiDAR point coverage. Conversely, strong occlusion reduces point returns and may cause missing detections. Since microscopic traffic simulators do not provide raw point clouds, we approximate the  LiDAR point coverage in a computationally efficient manner by counting how many azimuth rays can reach each surrounding vehicle from the ego AV.

\paragraph{Azimuthal ray discretization.}
We discretize the full $360^\circ$ field of view of the ego AV into $N_r$ evenly spaced azimuthal rays. Each ray corresponds to a fixed angular direction:
\begin{equation}
\theta_r = -\pi + \left(r+\tfrac{1}{2}\right)\Delta\theta,
\quad
\Delta\theta = \frac{2\pi}{N_r},
\quad
r = 0,\dots,N_r-1 .
\end{equation}
Each ray represents a narrow angular slice in the bird’s-eye-view (BEV) plane and serves as a surrogate for a group of LiDAR beams with similar azimuth angles.

\paragraph{Angular footprint of surrounding vehicles.}
For each surrounding vehicle $i$, we approximate its angular visibility by projecting its oriented BEV bounding box onto the azimuthal domain. Let the ego AV position be $\mathbf{p}_e=(x_e,y_e)$ and the four BEV corners of vehicle $i$ be $\{x_{i,m}, y_{i,m}\}_{m=1}^{4}$. The azimuth angle of each corner is computed as
\begin{equation}
\phi_{i,m}
=
\arctan2\!\left(y_{i,m}-y_e,\; x_{i,m}-x_e\right).
\end{equation}

We then determine the \emph{minimal continuous angular interval} $\mathcal{I}_i = [\phi_i^{\min}, \phi_i^{\max}]$ that covers all four corner angles, accounting for angular wrap-around at $\pm\pi$. A ray $r$ is said to intersect vehicle $i$ if $\theta_r \in \mathcal{I}_i$.

\paragraph{Ray-wise depth buffering (occlusion handling).}
Occlusion along each azimuthal ray is addressed using a depth-buffering rule. For a given ray direction $\theta_r$, multiple vehicles may intersect with the ray. Among them, only the nearest vehicle to the ego AV is considered visible along that ray, while all farther vehicles are occluded. Using the center-to-ego distance $d_i=\|\mathbf{p}_i-\mathbf{p}_e\|_2$ as a depth proxy, the per-slice ray-hit count for vehicle $i$ is defined as
\begin{equation}
RH_i^{(k)} = 
\sum_{r=0}^{N_r-1}
\mathbb{I}\!\left[
i = \arg\min_{j \in \mathcal{V}_r^{(k)}} d_j
\right],
\end{equation}
where $\mathcal{V}_r^{(k)}$ denotes the set of vehicles intersected by ray $r$ in height slice $k$ (defined below). Intuitively, $RH_i^{(k)}$ measures how many azimuthal rays directly reach vehicle $i$ without being blocked by closer vehicles.

\paragraph{Height-aware $k$-depth peeling.}
Purely 2D ray casting may overestimate occlusion, as real LiDAR systems emit beams that cover a wide vertical field of view (FOV) and vehicles can be partially visible in height. To address this limitation, HARC extends azimuthal ray casting with a \emph{height-aware $k$-depth peeling} strategy. Specifically, the vertical dimension is divided into $K$ height slices defined by thresholds
\begin{equation}
0 = z_0 < z_1 < \dots < z_K ,
\end{equation}
where slice $k$ corresponds to the interval $(z_{k-1}, z_k]$ with the height equals to $h_k = z_k - z_{k-1}$. In slice $k$, only vehicles whose height exceeds $z_{k-1}$ are considered active and participate in ray casting. The ray-wise depth-buffering process is repeated independently in each slice, yielding $RH_i^{(k)}$.

The final ray-hit feature is obtained by aggregating ray hits across all height slices using a height-weighted average:
\begin{equation}
RH_i
=
\frac{\sum_{k=1}^{K} h_k \, RH_i^{(k)}}
     {\sum_{k=1}^{K} h_k}.
\label{eq:harc_final}
\end{equation}
This formulation captures both lateral occlusion (angular blocking) and partial vertical visibility in a compact scalar feature. The resulting ray-hit value $RH_i$ is used purely as a node attribute in the graph representation, serving as a physically grounded visibility prior.

The overall flow for the ray-hit feature construction is illustrated in Algorithm~\ref{alg:harc}.

\begin{figure}[!htbp]
    \centering
    \includegraphics[scale=0.7]{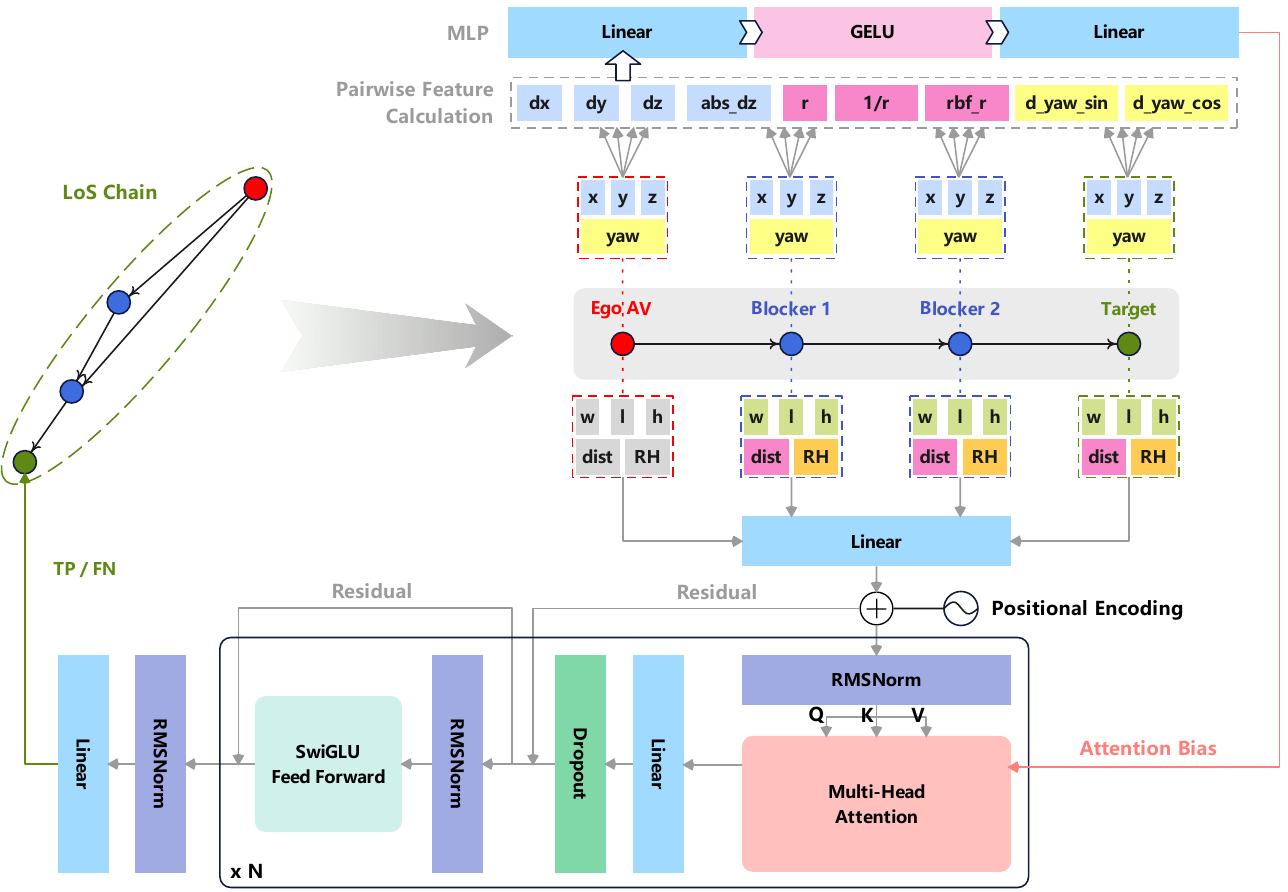}
    \caption{\textbf{Illustration of LoS-Graphormer Architecture}}
    \label{fig:LoSGraphormer}
\end{figure}

\subsection{Model Architecture}
To effectively model occlusion-aware perception under realistic traffic conditions, we design LoS-Graphormer, a transformer architecture tailored to Line-of-Sight (LoS) structured inputs. As illustrated in Fig.~\ref{fig:LoSGraphormer}, LoS-Graphormer operates on LoS chains derived from the RM-LoS graph, where each chain represents an ordered sequence of vehicles along the sight of AV to the target vehicle.

The network consists of three key components. First, node-level feature embeddings are constructed for each vehicle in the LoS chain using vehicle positions and headings. Second, a pairwise feature module computes relative geometric relationships between vehicles in the chain and projects them through a lightweight MLP to generate an attention bias, which injects physically meaningful spatial priors into the transformer. Third, a Transformer encoder block, composed of multi-head self-attention, normalization layers, residual connections, and a feed-forward network, performs contextual reasoning along each LoS chain.

By combining structured LoS chains with geometry-aware attention modulation, the LoS-Graphormer focuses computation on physically relevant vehicle interactions while maintaining architectural simplicity and scalability. The network outputs a visibility prediction for each target vehicle based on the aggregated chain representation, enabling realistic surrogate LiDAR detection without relying on raw point-cloud data.

\section*{Data availability}
The datasets generated for MIDAR training and evaluation are available at \url{https://github.com/Purdue-CART-Lab/MIDAR/tree/main/data}. 
The raw LiDAR point cloud data collected from the SUMO--CARLA co-simulation platform are available from the corresponding author upon reasonable request. 
The nuScenes dataset is publicly available at \url{https://www.nuscenes.org/nuscenes}.

\section*{Code availability}
The code used to train and evaluate the models in this study is available at \url{https://github.com/Purdue-CART-Lab/MIDAR}.

\section*{Acknowledgments}
This research was partially funded by the the National Science Foundation (CMMI 2339753) and U.S. Department of Transportation (USDOT) Region 5 University Transportation Center: Center for Connected and Automated Transportation (CCAT) (69A3551747105). Any opinions, findings, conclusions, or recommendations expressed in this material are those of the authors and do not necessarily reflect the official policy or position of the U.S. government.


\section*{Author contribution}
The authors confirm contribution to the paper as follows: study conception and design: Zhu and Feng; data collection: Zhu; analysis and interpretation of results: Zhu and Feng; manuscript preparation: Zhu and Feng. Both authors reviewed the results and approved the final version of the manuscript.


\section*{Competing interests}

The authors declare no competing interests.




\bibliographystyle{naturemag}  
\bibliography{refs}          

@article{feng2015a,
   author = {Feng, Yiheng and Head, K. Larry and Khoshmagham, Shayan and Zamanipour, Mehdi},
   title = {A real-time adaptive signal control in a connected vehicle environment},
   journal = {Transportation Research Part C: Emerging Technologies},
   volume = {55},
   pages = {460-473},
   ISSN = {0968090X},
   DOI = {10.1016/j.trc.2015.01.007},
   year = {2015},
   type = {Journal Article}
}

@article{garey1976graph,
   author = {Garey, Michael and Johnson, David and So, Hing},
   title = {An application of graph coloring to printed circuit testing},
   journal = {IEEE Transactions on circuits and systems},
   volume = {23},
   number = {10},
   pages = {591-599},
   ISSN = {0098-4094},
   year = {1976},
   type = {Journal Article}
}

@article{guo2021mixed,
   author = {Guo, Qiangqiang and Ban, Xuegang Jeff and Aziz, HM Abdul},
   title = {Mixed traffic flow of human driven vehicles and automated vehicles on dynamic transportation networks},
   journal = {Transportation research part C: emerging technologies},
   volume = {128},
   pages = {103159},
   ISSN = {0968-090X},
   year = {2021},
   type = {Journal Article}
}

@misc{LTAdatamall,
   title =  {{LTA Datamall}},
   url = {https://datamall.lta.gov.sg},
   year = {n.d.},
   type = {Web Page}
}

@article{li2021cooperative,
   author = {Li, Tao and Han, Xu and Ma, Jiaqi},
   title = {Cooperative perception for estimating and predicting microscopic traffic states to manage connected and automated traffic},
   journal = {IEEE Transactions on Intelligent Transportation Systems},
   volume = {23},
   number = {8},
   pages = {13694-13707},
   ISSN = {1524-9050},
   year = {2021},
   type = {Journal Article}
}

@article{li2024a,
   author = {Li, Wangzhi and Zhu, Tianheng and Feng, Yiheng},
   title = {A cooperative perception based adaptive signal control under early deployment of connected and automated vehicles},
   journal = {Transportation Research Part C: Emerging Technologies},
   volume = {169},
   pages = {104860},
   ISSN = {0968-090X},
   DOI = {https://doi.org/10.1016/j.trc.2024.104860},
   url = {https://www.sciencedirect.com/science/article/pii/S0968090X24003814},
   year = {2024},
   type = {Journal Article}
}

@inproceedings{Lope2018sumo,
   author = {Lopez, Pablo Alvarez and Behrisch, Michael and Bieker-Walz, Laura and Erdmann, Jakob and Flötteröd, Yun-Pang and Hilbrich, Robert and Lücken, Leonhard and Rummel, Johannes and Wagner, Peter and Wießner, Evamarie},
   title = {Microscopic traffic simulation using sumo},
   booktitle = {2018 21st international conference on intelligent transportation systems (ITSC)},
   publisher = {IEEE},
   pages = {2575-2582},
   ISBN = {1728103231},
   type = {Conference Proceedings}
}

@article{loshchilov2017decoupled,
   author = {Loshchilov, Ilya and Hutter, Frank},
   title = {Decoupled weight decay regularization},
   journal = {arXiv preprint arXiv:1711.05101},
   year = {2017},
   type = {Journal Article}
}

@misc{openpcdet,
   title = {{OpenPCDet}},
   url = {https://github.com/open-mmlab/OpenPCDet},
   year = {2020},
   type = {Web Page}
}

@inproceedings{LGSVL,
   author = {Rong, Guodong and Shin, Byung Hyun and Tabatabaee, Hadi and Lu, Qiang and Lemke, Steve and Možeiko, Mārtiņš and Boise, Eric and Uhm, Geehoon and Gerow, Mark and Mehta, Shalin},
   title = {Lgsvl simulator: A high fidelity simulator for autonomous driving},
   booktitle = {2020 IEEE 23rd International conference on intelligent transportation systems (ITSC)},
   publisher = {IEEE},
   pages = {1-6},
   ISBN = {1728141494},
   type = {Conference Proceedings}
}

@inproceedings{AirSim,
   author = {Shah, Shital and Dey, Debadeepta and Lovett, Chris and Kapoor, Ashish},
   title = {Airsim: High-fidelity visual and physical simulation for autonomous vehicles},
   booktitle = {Field and service robotics: Results of the 11th international conference},
   publisher = {Springer},
   pages = {621-635},
   type = {Conference Proceedings}
}

@article{wei2021a,
   author = {Wei, Lei and Wang, Yunpeng and Chen, Peng},
   title = {A Particle Filter-Based Approach for Vehicle Trajectory Reconstruction Using Sparse Probe Data},
   journal = {IEEE Transactions on Intelligent Transportation Systems},
   volume = {22},
   number = {5},
   pages = {2878-2890},
   ISSN = {1524-9050
1558-0016},
   DOI = {10.1109/tits.2020.2976671},
   year = {2021},
   type = {Journal Article}
}

@inproceedings{yin2021center,
  title={Center-based 3d object detection and tracking},
  author={Yin, Tianwei and Zhou, Xingyi and Krahenbuhl, Philipp},
  booktitle={Proceedings of the IEEE/CVF conference on computer vision and pattern recognition},
  pages={11784--11793},
  year={2021}
}

@article{zhang2024vehicle,
   author = {Zhang, Cong and Feng, Yiheng},
   title = {Vehicle trajectory reconstruction for freeway traffic considering lane changing behaviors},
   journal = {Journal of Intelligent Transportation Systems},
   pages = {1-16},
   ISSN = {1547-2450
1547-2442},
   DOI = {10.1080/15472450.2024.2307031},
   year = {2024},
   type = {Journal Article}
}

@article{zhang2018mitigating,
   author = {Zhang, Kenan and Nie, Yu Marco},
   title = {Mitigating the impact of selfish routing: An optimal-ratio control scheme (ORCS) inspired by autonomous driving},
   journal = {Transportation Research Part C: Emerging Technologies},
   volume = {87},
   pages = {75-90},
   ISSN = {0968-090X},
   year = {2018},
   type = {Journal Article}
}

@article{chen2021graph,
  title={Graph neural network and reinforcement learning for multi-agent cooperative control of connected autonomous vehicles},
  author={Chen, Sikai and Dong, Jiqian and Ha, Paul and Li, Yujie and Labi, Samuel},
  journal={Computer-Aided Civil and Infrastructure Engineering},
  volume={36},
  number={7},
  pages={838--857},
  year={2021},
  publisher={Wiley Online Library}
}

@article{liu2024reinforcement,
  title={Reinforcement learning-based multi-lane cooperative control for on-ramp merging in mixed-autonomy traffic},
  author={Liu, Lin and Li, Xiaoxuan and Li, Yongfu and Li, Jingxiang and Liu, Zhongyang},
  journal={IEEE Internet of Things Journal},
  year={2024},
  publisher={IEEE}
}

@article{chen2024macro,
  title={A macro-micro approach to reconstructing vehicle trajectories on multi-lane freeways with lane changing},
  author={Chen, Xuejian and Qin, Guoyang and Seo, Toru and Yin, Juyuan and Tian, Ye and Sun, Jian},
  journal={Transportation research part C: emerging technologies},
  volume={160},
  pages={104534},
  year={2024},
  publisher={Elsevier}
}

@inproceedings{CARLA,
  title={CARLA: An open urban driving simulator},
  author={Dosovitskiy, Alexey and Ros, German and Codevilla, Felipe and Lopez, Antonio and Koltun, Vladlen},
  booktitle={Conference on robot learning},
  pages={1--16},
  year={2017},
  organization={PMLR}
}

@misc{AWSIM,
  title        = {AWSIM: Autonomous Driving Simulator},
  author       = {{Autoware Foundation}},
  year         = {2022},
  howpublished = {\url{https://github.com/tier4/AWSIM}},
  note         = {Accessed: 2025-01-01}
}

@misc{vissim,
  title        = {PTV Vissim: Traffic Simulation Software},
  author       = {{PTV Group}},
  year         = {2023},
  howpublished = {\url{https://www.ptvgroup.com/en/solutions/products/ptv-vissim/}},
  note         = {Accessed: 2025-01-01}
}

@inproceedings{xu2021opencda,
  title={Opencda: an open cooperative driving automation framework integrated with co-simulation},
  author={Xu, Runsheng and Guo, Yi and Han, Xu and Xia, Xin and Xiang, Hao and Ma, Jiaqi},
  booktitle={2021 IEEE International Intelligent Transportation Systems Conference (ITSC)},
  pages={1155--1162},
  year={2021},
  organization={IEEE}
}

@inproceedings{WOD,
  title={Scalability in perception for autonomous driving: Waymo open dataset},
  author={Sun, Pei and Kretzschmar, Henrik and Dotiwalla, Xerxes and Chouard, Aurelien and Patnaik, Vijaysai and Tsui, Paul and Guo, James and Zhou, Yin and Chai, Yuning and Caine, Benjamin and others},
  booktitle={Proceedings of the IEEE/CVF conference on computer vision and pattern recognition},
  pages={2446--2454},
  year={2020}
}

@inproceedings{caesar2020nuscenes,
  title={nuscenes: A multimodal dataset for autonomous driving},
  author={Caesar, Holger and Bankiti, Varun and Lang, Alex H and Vora, Sourabh and Liong, Venice Erin and Xu, Qiang and Krishnan, Anush and Pan, Yu and Baldan, Giancarlo and Beijbom, Oscar},
  booktitle={Proceedings of the IEEE/CVF conference on computer vision and pattern recognition},
  pages={11621--11631},
  year={2020}
}

@misc{CARLA_SUMO_OAC,
  title        = {CARLA--SUMO Co-Simulation Framework},
  author       = {{NSF Open Autonomous Corridor (OAC) Project}},
  year         = {2023},
  howpublished = {\url{https://nsf-oac.readthedocs.io/en/latest/index.html}},
  note         = {Accessed: 2025-01-01}
}

@article{Chen01012021,
    author = {Peng Chen and Lei Wei and Fangfang Meng and Nan Zheng},
    title = {Vehicle trajectory reconstruction for signalized intersections: A hybrid approach integrating Kalman Filtering and variational theory},
    journal = {Transportmetrica B: Transport Dynamics},
    volume = {9},
    number = {1},
    pages = {22--41},
    year = {2021},
    publisher = {Taylor \& Francis},
    doi = {10.1080/21680566.2020.1781707},
    URL = { 
            https://doi.org/10.1080/21680566.2020.1781707
    }
}

@article{zhu2026cptrajrecon,
  title = {Integrated optimization for vehicle trajectory reconstruction under cooperative perception environment},
  author = {Tianheng Zhu and Wangzhi Li and Yiheng Feng},
  journal = {Transportation Research Part C: Emerging Technologies},
  volume = {184},
  pages = {105522},
  year = {2026},
  publisher={Elsevier}
}

@article{ye2019evaluating,
  title={Evaluating the impact of connected and autonomous vehicles on traffic safety},
  author={Ye, Lanhang and Yamamoto, Toshiyuki},
  journal={Physica A: Statistical Mechanics and its Applications},
  volume={526},
  pages={121009},
  year={2019},
  publisher={Elsevier}
}

@article{garg2023can,
  title={Can connected autonomous vehicles improve mixed traffic safety without compromising efficiency in realistic scenarios?},
  author={Garg, Mohit and Bouroche, M{\'e}lanie},
  journal={IEEE Transactions on Intelligent Transportation Systems},
  volume={24},
  number={6},
  pages={6674--6689},
  year={2023},
  publisher={IEEE}
}

@article{peng2021connected,
  title={Connected autonomous vehicles for improving mixed traffic efficiency in unsignalized intersections with deep reinforcement learning},
  author={Peng, Bile and Keskin, Musa Furkan and Kulcs{\'a}r, Bal{\'a}zs and Wymeersch, Henk},
  journal={Communications in Transportation Research},
  volume={1},
  pages={100017},
  year={2021},
  publisher={Elsevier}
}

@article{luo2024stabilizing,
  title={Stabilizing traffic flow by autonomous vehicles: Stability analysis and implementation considerations},
  author={Luo, Lihua and Liu, Yi and Feng, Yiheng and Liu, Henry X and Ge, Ying-En},
  journal={Transportation research part C: emerging technologies},
  volume={158},
  pages={104449},
  year={2024},
  publisher={Elsevier}
}

@article{wang2023general,
  title={A general approach to smoothing nonlinear mixed traffic via control of autonomous vehicles},
  author={Wang, Shian and Shang, Mingfeng and Levin, Michael W and Stern, Raphael},
  journal={Transportation Research Part C: Emerging Technologies},
  volume={146},
  pages={103967},
  year={2023},
  publisher={Elsevier}
}

@techreport{bayartsengel2024enhancing,
  title={Enhancing Vulnerable Road User Safety at Signalized Intersections Through Cooperative Perception and Driving Automation},
  author={Bayartsengel, Misheel and Soleimaniamiri, Saeid and Huang, Zhitong and Wang, Qinzheng and Racha, Sujith and others},
  year={2024},
  institution={United States. Federal Highway Administration. Office of Safety and~…}
}

@article{krajzewicz2012recent,
  title={Recent development and applications of SUMO-Simulation of Urban MObility},
  author={Krajzewicz, Daniel and Erdmann, Jakob and Behrisch, Michael and Bieker, Laura and others},
  journal={International journal on advances in systems and measurements},
  volume={5},
  number={3\&4},
  pages={128--138},
  year={2012}
}

@article{geiger2013vision,
  title={Vision meets robotics: The kitti dataset},
  author={Geiger, Andreas and Lenz, Philip and Stiller, Christoph and Urtasun, Raquel},
  journal={The international journal of robotics research},
  volume={32},
  number={11},
  pages={1231--1237},
  year={2013},
  publisher={Sage Publications Sage UK: London, England}
}

@article{kuhn1955hungarian,
  title={The Hungarian method for the assignment problem},
  author={Kuhn, Harold W},
  journal={Naval research logistics quarterly},
  volume={2},
  number={1-2},
  pages={83--97},
  year={1955},
  publisher={Wiley Online Library}
}

@article{cao2022analytical,
  title={An analytical model for quantifying the efficiency of traffic-data collection using instrumented vehicles},
  author={Cao, Peng and Xiong, Zhiqiang and Liu, Xiaobo},
  journal={Transportation research part C: emerging technologies},
  volume={136},
  pages={103558},
  year={2022},
  publisher={Elsevier}
}

\appendix

\section{Algorithms Pseudo Code}
\begin{algorithm}[H]
\caption{Frame-wise TP/FN/FP Labeling via IoU and Hungarian Matching}
\label{alg:iou_hungarian_labeling}
\begin{algorithmic}[1]
\State \textbf{Input:} GT boxes per frame $\{\mathcal{G}_t\}$, predictions per frame $\{\mathcal{P}_t\}$, confidence threshold $\tau_c{=}0.3$, IoU threshold $\tau_{iou}{=}0.5$
\State \textbf{Output:} Labels for GT boxes (TP/FN) and predicted boxes (TP/FP)
\For{timestamp $t$ in all frames}
    \State $\mathcal{G} \gets \mathcal{G}_t$; \quad $\mathcal{P} \gets \mathcal{P}_t$
    \State $\mathcal{P} \gets \{p \in \mathcal{P}\mid \mathrm{conf}(p)\ge \tau_c\}$ \Comment{filter by confidence}
    \State Initialize all $g\in\mathcal{G}$ as FN; initialize all $p\in\mathcal{P}$ as FP
    \For{each class $c$}
        \State $\mathcal{G}^c \gets \{g\in\mathcal{G}\mid \mathrm{class}(g)=c\}$
        \State $\mathcal{P}^c \gets \{p\in\mathcal{P}\mid \mathrm{class}(p)=c\}$
        \If{$|\mathcal{G}^c|=0$ \textbf{or} $|\mathcal{P}^c|=0$}
            \State \textbf{continue}
        \EndIf
        \State Compute IoU matrix $M \in \mathbb{R}^{|\mathcal{G}^c|\times|\mathcal{P}^c|}$ where $M_{ij}=\mathrm{IoU3D}(g_i,p_j)$
        \State Construct cost matrix $C$ with entries $C_{ij}=-M_{ij}$
        \State $\pi \gets \mathrm{Hungarian}(C)$ \Comment{optimal one-to-one assignment}
        \For{each matched pair $(i,j)\in \pi$}
            \If{$M_{ij}\ge \tau_{iou}$}
                \State Label $g_i$ as TP; label $p_j$ as TP
            \EndIf
        \EndFor
    \EndFor
\EndFor
\end{algorithmic}
\end{algorithm}

\begin{algorithm}[H]
\caption{Height-Aware Azimuthal Ray Casting (HARC) for Ray-Hit}
\label{alg:harc}
\begin{algorithmic}[1]
\State \textbf{Initialize:} Number of rays, height thresholds
\State \textbf{Inputs:} Ego pose $\mathbf{p}_e$, surrounding vehicles $\{\mathbf{p}_i, H_i, \text{bbox}_i\}$
\State \textbf{Output:} Ray-hit feature $\{RH_i\}$

\State \textbf{// 1. Ray Discretization}
\State Uniformly discretize $360^\circ$ FOV into $N_r$ azimuthal rays $\{\theta_r\}$

\State \textbf{// 2. Angular Projection}
\For{each vehicle $i$}
    \State Compute corner azimuth angles from bounding box
    \State Determine minimal angular interval $\mathcal{I}_i$
    \State Compute distance $d_i$ to ego
\EndFor

\State \textbf{// 3. Height-Aware Ray Casting}
\For{height slice $k = 1,\dots,K$}
    \State Activate vehicles with height $H_i > z_{k-1}$
    \State Initialize $RH_i^{(k)} \gets 0$
    \For{ray $r = 1,\dots,N_r$}
        \State Find vehicles intersecting ray $\theta_r$
        \State Assign ray to nearest vehicle (depth buffering)
        \State Increment corresponding $RH_i^{(k)}$
    \EndFor
\EndFor

\State \textbf{// 4. Aggregation}
\For{each vehicle $i$}
    \State Compute weighted ray-hit:
    \State \quad $RH_i \gets \dfrac{\sum_k h_k RH_i^{(k)}}{\sum_k h_k}$
\EndFor

\State \Return $\{RH_i\}$
\end{algorithmic}
\end{algorithm}

\section{LiDAR Specifications for Training Data Collection} 
\label{appendix:lidarspec}

\begin{table}[H]
\centering
\caption{Specifications of the Velodyne HDL-32E LiDAR used in nuScenes.}
\label{tab:hdl32e}
\begin{tabular}{ll}
\hline
\textbf{Specification} & \textbf{Value} \\
\hline
Capture frequency      & 20 Hz \\
Number of beams        & 32 \\
Channels               & 32 \\
Horizontal FOV         & 360$^\circ$ \\
Vertical FOV           & +10$^\circ$ to $-30^\circ$ \\
Maximum range          & 80-100 m \\
Point rate             & Up to $\sim$1.39 million points/s \\
Range accuracy         & $\pm$2 cm \\
\hline
\end{tabular}
\end{table}

\begin{table}[H]
\centering
\caption{Specifications of the simulated LiDAR sensor used in CARLA.}
\label{tab:carla_LiDAR}
\begin{tabular}{ll}
\hline
\textbf{Specification} & \textbf{Value} \\
\hline
Capture frequency        & 20 Hz \\
Number of beams          & 64 \\
Channels                 & 64 \\
Horizontal FOV           & 360$^\circ$ \\
Vertical FOV             & +10$^\circ$ to $-30^\circ$ \\
Maximum range            & 80 m \\
Point rate               &  Up to $\sim$1.28 million points/s \\
Range accuracy           & Simulator-defined (noise-free) \\
\hline
\end{tabular}
\end{table}

\section{Process-Level Resource Usage Breakdown}
\label{app:resource_breakdown}

\begin{table}[H]
\centering
\caption{Average process-level resource usage at 10\% AV MPR.}
\label{tab:resource_breakdown_10}
\setlength{\tabcolsep}{5pt}
\begin{tabular}{l l cc}
\toprule
Method & Process & Avg CPU Mem (GB) & Avg GPU Mem (GB) \\
\midrule
MIDAR & Python (MIDAR Pipeline) & 0.96 & 0.48 \\
MIDAR & SUMO   & 0.11 & 0.00 \\
\midrule
Conventional & Python (Centerpoint) & 4.87 & 18.22 \\
Conventional & CARLA  & 3.03 & 3.05 \\
Conventional & SUMO   & 0.11 & 0.00 \\
\bottomrule
\end{tabular}
\end{table}

\end{document}